\documentclass[accepted]{article}

\usepackage{microtype}
\usepackage{graphicx}
\usepackage{subcaption}
\usepackage{booktabs} 
\usepackage{rays_defs_18}
\usepackage{hyperref}

\newtheorem{lemma}{Lemma}

\newtheorem{theorem}{Theorem}

\newcommand\median{\mathrm{median}}

\usepackage{mlsys2022}

\mlsystitlerunning{Federated Multiple Label Hashing (FedMLH): Communication Efficient Federated Learning on Extreme Classification Tasks}

\begin{document}

\twocolumn[
\mlsystitle{Federated Multiple Label Hashing (FedMLH): Communication Efficient Federated Learning on Extreme Classification Tasks}



\mlsyssetsymbol{equal}{*}

\begin{mlsysauthorlist}
\mlsysauthor{Zhenwei Dai}{equal,stat}
\mlsysauthor{Chen Dun}{equal,comp}
\mlsysauthor{Yuxin Tang}{equal,comp}
\mlsysauthor{Anastasios Kyrillidis}{comp}
\mlsysauthor{Anshumali Shrivastava}{comp}
\end{mlsysauthorlist}

\mlsysaffiliation{stat}{Department of Statistics, Rice University, Texas, USA}
\mlsysaffiliation{comp}{Department of Computer Science, Rice University, Texas, USA}

\mlsyscorrespondingauthor{Anshumali Shrivastava}{anshumali@rice.edu}

\mlsyskeywords{Machine Learning, MLSys}

\vskip 0.3in

\begin{abstract}
Federated learning enables many local devices to train a deep learning model jointly without sharing the local data. Currently, most of federated training schemes learns a global model by averaging the parameters of local models. 
However, most of these training schemes suffer from high communication cost resulted from transmitting full local model parameters. Moreover, directly averaging model parameters leads to a significant performance degradation, due to the class-imbalanced non-iid data on different devices. Especially for the real life federated learning tasks involving extreme classification, (1) communication becomes the main bottleneck since the model size increases proportionally to the number of output classes; (2) extreme classification (such as user recommendation) normally have extremely imbalanced classes and heterogeneous data on different devices. To overcome this problem, we propose federated multiple label hashing (FedMLH), which leverages label hashing to simultaneously reduce the model size (up to \textbf{3.40}$\times$ decrease) with communication cost (up to \textbf{18.75}$\times$ decrease) and achieves significant better accuracy (up to \textbf{35.5\%} relative accuracy improvement) and faster convergence rate (up to \textbf{5.5}$\times$ increase) for free on the federated extreme classification tasks compared to federated average algorithm.
\end{abstract}
]



\printAffiliationsAndNotice{\mlsysEqualContribution} 

\section{Introduction}
\label{introduction}
As a lot of modern edge devices, like smart phones and IoT devices, keep generating massive data, learning a model locally on a large number of devices has become more and more important for machine learning.
With growing computation power of the edge devices and higher requirement of data privacy, federated learning (FL) has become one of the important domains in large scale machine learning. Different from the traditional centralized learning, federated learning does not store the data and train machine learning models on a central server. Instead, the data are saved on the local clients without sharing with others, and most of the computations are completed locally. In detail, FL lets the local clients learn a local model using its user generated data. To train a global model that can be generalized to different users, FL uses a central server to collect the local model parameters and aggregate them into a global model periodically.

FL has achieved a great success in many different types of machine learning tasks. In this project, we focused on FL on extreme classification tasks (federated extreme classification). 
Extreme classification task requires to predict the labels of a large number of different classes~\cite{choromanska2014logarithmic, prabhu2018parabel, hsu2009multi, medini2019extreme}. 
Due to the more and more strict privacy regulations, federated extreme classification has found a wide range of application scenarios. 
For example, many social media companies are interested in training NLP models (most NLP models involve predicting word from extreme large vocabulary) based on user generated content 
for sentimental analysis, inappropriate language detection or text auto completion. However, due to the privacy regulations like GDPR, it becomes difficult to share user data on social media across boarders (like the investigations on Facebook and WhatsApp user data policy~\cite{GDPR}). Hence, FL is a good solution to learn a good NLP model without sharing users' data from different countries~\cite{lin2021fednlp}. Another example is training product/advertisement recommendation systems on e-commerce platforms based on users' data. The recommendation systems also involve hundreds thousands of different items as output labels. Due to the similar dilemma faced by the social media platforms, we may also need to train a recommendation system using FL algorithms~\cite{yang2020federated}.

Different from the traditional FL tasks, federated extreme classification faces some unique challenges. First, federated extreme classification usually has highly imbalanced class distribution (figure~\ref{fig:label_freq}). Most of the classes only have a few positive instances despite of the large sample size. Second, due to large number of output classes, extreme classification models has memory blow up in the last fully connected layer. Hence, the communication cost of each synchronization round is huge. Moreover, FL usually requires more training epochs before converges, which adds extra burden to the model communications.
Third, for the non-iid partitioned local datasets,  in federated extreme classification tasks, the class distributions are divergent between different local datasets. A local dataset may have a lot of positive instances in class $j$ while another local dataset may only contain negative instances of class $j$. Especially when the FL task has a large number of classes (like federated extreme classification tasks), this divergence is even more obvious compared to the FL tasks with a small number of classes (see theorem~\ref{theorem:divergence}). Previous research suggests that the class distribution divergence significantly hurt the generalization performance of the global model~\cite{zhao2018federated,karimireddy2020scaffold}.

However, these challenges have not been well addressed by the current FL algorithms. For example, as the most popular FL algorithm, federated average algorithm (FedAvg)~\cite{mcmahan2017communication} learns a global model by periodically transmitting and averaging the parameters of local models trained on the local data. However, FedAvg does not overcome the above challenges: (1) large data heterogeneity significantly degrades the performance of FedAvg and slows down the convergence ~\cite{sahu2018convergence, karimireddy2020scaffold}; (2) Though models are only periodically synchronized, transmitting all local full models still results in high communication cost. Thus, naively applying FedAvg on extreme classification will have large performance degradation, slow convergence and extremely high communication cost.

\paragraph{Our Contributions:} In this paper, we propose a novel method, Federated Multiple Label Hashing (FedMLH), for federated extreme classification tasks. Compared to FedAvg algorithm, FedMLH significantly reduces the communication cost, improves the model accuracy and speeds up the training. Moreover, FedMLH compresses the model size, adjusts the imbalanced class size and also reduces the class distribution divergence between different local clients. We evaluate FedMLH on four different extreme classification datasets, and FedMLH consistently demonstrates much better performance. FedMLH reduces the communication cost by up to $\textbf{18.75}\times$, improves the absolute prediction accuracy by up to \textbf{9\%} and relative accuracy improvement by up to \textbf{35.5\%}, speeds up the convergence rate by up to $\textbf{5.5}\times$, which is a remarkable improvement. 

We also provide theoretical analysis to show that FedMLH relieves problem of imbalanced class size and divergent class distribution between clients.

\section{Problem Statement}

First, we define the setup of federated learning in our paper. Let $(\vx, \vy)$ be the input features and labels pair, where feature vector $\vx \in \reals^d$ and label $\vy = \{0, 1\}^p$. Assume we have $K$ local devices, and each device $k$ generates its own dataset $D_k$ for $k=1,2,\ldots,K$. Let $n_k$ be sample size of $D_k$, $n_k = |D_k|$, and $N$ is the total number of samples on all the local devices. A local model $\vw^k$ on device $k$ is trained using dataset $D_k$ and the corresponding empirical loss function is defined
$L(\vw^k|D_k) = \frac{1}{n_k} \sum_{\vx_i, \vy_i \in D_k} \ell(\vx_i, \vy_i | \vw^k)$,
where $\ell(\vx_i, \vy_i|\vw^k)$ is the loss function for sample $(\vx_i, \vy_i)$ under parameter $\vw^k$. A global model $\vw$ is learned from the local models $\vw_k$. The performance of the federated learning model is measured using the global model $\vw$ on the testing set.

\paragraph{Non-iid local data distribution} In many FL tasks, the local datasets, $D_k$ (data on $k$-th client), are not following the same distribution, i.e., for $(\vx, \vy) \in D_k$, $(\vx, \vy) \sim F_k$ but $F_k$ varies across different local clients. 
Especially for extreme classification tasks, since the number of classes is huge, the local data distribution $F_k$ is even more divergent. Hence, it is more practical to assume the $F_k$ are non-iid. In the experiment section, we design a data partition mechanism to ensure $D_k$ following totally different distributions. 

\section{Background}
We start by introducing the FedAvg algorithm and how to use Count Sketch to compress the data. 

\subsection{FedAvg algorithm}
FedAvg algorithm learns a global model by directly averaging the local parameters at synchronization. Assume each global iteration involves $M$ epochs. At the beginning of a global iteration $t$, the central server randomly picks a subset $S_{t}$ of $K^{'}$ local devices and broadcast the global weight $\vw_{(t)}$ to the selected local devices. 
Then, on the selected local device $k$, it updates the model parameters from $\vw_{(t)}$ to $\vw^{k}_{(t)}$ using the local loss function, $f_{k}$. At synchronization, the global model is updated by averaging $\vw^k_{(t+1)}$, $\vw_{(t+1)} = \sum_{k \in S_{t}}\frac{n_k}{N} \vw^{k}_{(t+1)}$. The FedAvg algorithm runs for $T$ global iterations to learn a global model.

\subsection{Count sketch}
Count sketch is a probabilistic data structure widely used to recover the heavy hitters. A count sketch consists of $K$ hash tables and each hash table has $R$ buckets. A vector $\vx = (\vx_1, \vx_2, \ldots, \vx_p) \in \reals^p$ is mapped into the hash tables using $K$ independent hash functions. Let $\mM \in \reals^{K \times R}$ be the matrix storing the values in the count sketches, and let $h_1, h_2, \ldots, h_K$ be independent uniform hash functions $h_j \colon \{1,2,\ldots,p\} \to \{1,2,\ldots,R\}$. In addition, count sketch uses sign hash functions $s_j \colon \{1,2,\ldots,p\} \to \{+1, -1\}$ to map the components of the vectors randomly to $\{+1, -1\}$ (Algorithm~\ref{alg:algo1}, line 3 to 5). 
To retrieve the estimate of $X_i$, again, count sketch computes the hashing locations, $(h_1(i), h_2(i), \ldots,h_K(i))$ and retrieves the values stored in the corresponding buckets. Then, take the median of the retrieved values as the estimate, $\hat{\mu}_{i} = \median_{j} \mM_{j, h_j(i)} \cdot s_{j}(i)$. We may also take the mean of $\mM_{j, h_j(i)} \cdot s_j(i)$ instead of taking median since by the law of large numbers, mean also gives a good central estimate.

\begin{algorithm}[H]
\caption{Count Sketch Algorithm}
\label{alg:algo1}
\begin{algorithmic}[1]
  \STATE \textbf{Input}: $\vx \in \reals^p$, $K$ independent uniform hash functions $h_1, h_2,\ldots, h_K$, and sign functions $s_1, s_2, \ldots, s_K$
  \STATE Initialize entries of hash table array $\mM \in \reals^{K \times R}$ to zero
  \FOR{$i = 1,2,\ldots, p$}
  \STATE Insertion: update $\mM_{j, h_j(i)} += \vx_i \cdot s_j(i)$ for $j=1,2,\ldots,K$
  \ENDFOR
  \STATE \textbf{Retrieval:} estimate of $\vx_i$, $\hat{\mu}_i = \median_j \mM_{j, h_j(i)} \cdot s_j(i)$
\end{algorithmic}
\end{algorithm}

\vspace{-0.4cm}
\section{Our proposal: Federated Multiple Label Hashing (FedMLH)}

In the introduction section, we have discussed the challenges of federated extreme classification. To address these challenges, we propose FedMLH, which reduces the communication cost and adjust the non-iid class distributions.

\begin{figure*}[ht]
\centering
\subfloat[\label{summary_fig}]{\includegraphics[width=0.7\textwidth]{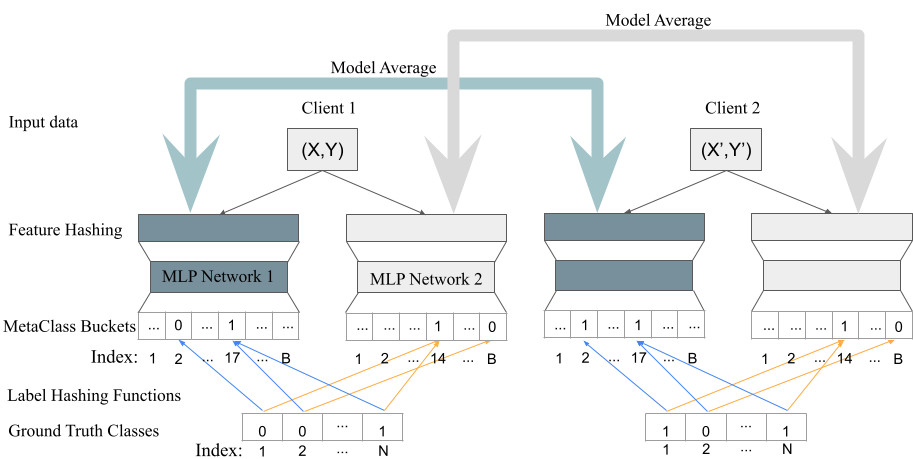}} \hfill
\subfloat[\label{summary_fig_2}]{\includegraphics[width=0.27\textwidth]{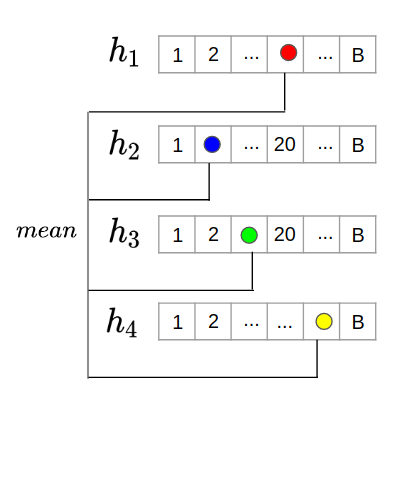}} 
\caption{(a) An example of FedMLH with two clients and each client has two different hash tables and predictor networks. In each hash table, $p$ classes are hashed into $B$ metaclass buckets. (b) FedMLH will merge the output probability for each hash table with $h_j, j=1,2,3,4$.}
\end{figure*}

\paragraph{Compress the last layer}
Due to the large number of output classes, the last fully connected layer of extreme classification models have a huge amount of parameters, which becomes the main communication bottleneck of the federated extreme classification. FedMLH leverages the idea of count sketch and hashes every class into $R$ independent hash tables (figure~\ref{summary_fig}). For sample $n$ and hash table $j$, its corresponding label of $i$-th bucket, $z^{j,k}_{n,i}$, is equal to the union of the class labels that are hashed into the same bucket (line 4-7, algorithm~\ref{alg:algo2}). Then $\vz^{j,k} \in \reals^{n_k \times B}$ contains the $j$-th hash table's bucket labels of all samples on client $k$. To compress the size of the last fully connect layer, we set the number of buckets ($B$) in each hash table to be much smaller than the number of classes. 

During the training, we use the bucket labels as the training target. Similar to the idea of count sketch, during the inference, to estimate the log-probability of a class $j$, we just go back to $R$ buckets that class $j$ is hashed into, and take the mean of the buckets' log-likelihoods as the log-likelihood of class $j$ (figure~\ref{summary_fig_2}).
Since the $R$ hash tables are independent to each other, for each hash table, we train an independent model (denoted as \textbf{``sub-model''}) to learn the corresponding bucket labels. Thus, FedMLH trains $R$ sub-models simultaneously. 

\paragraph{Training and model synchronization}
First, the central server generates hash table size $B$ and $R$ independent 2-universal hash functions, $h_1, h_2, \ldots, h_R$, and then broadcast to the local clients (line 3, algorithm~\ref{alg:algo2}). On the local clients, the output classes are hashed into $R$ hash tables using $h_i$. We also initialize $R$ independent sub-models which use the bucket labels as the training target. Let $\vw^{j,k}$ be the parameter of the $j$-th sub-model on the $k$-th client.

\begin{algorithm}[H]
\caption{FedMLH}
\label{alg:algo2}
\begin{algorithmic}[1]
    \STATE \textbf{Input:} Number of selected clients $S$, hash function number $R$, hash table size $B$, training data on $k$-th client $(\vx^{k}, \vy^{k})$.
    \STATE \textbf{On central server:} Generate $R$ uniform hash functions, $h_1, h_2, \ldots, h_R$, on the server, where $h_j: \{0,1,\ldots,p-1\} \rightarrow \{0,1,\ldots, B-1\}$ 
    \STATE Broadcast the $R$ hash functions and hash table size $B$ to each local client
    \newline
    \STATE \textbf{Label hashing on client $k$}, $k=1,2,\ldots,K$
    \FOR {$i=1,2,\ldots,B$, $j=1,2,\ldots,R$, $n=1,2,\ldots,n_k$}
    \STATE label of $i$-th bucket, $z^{j,k}_{n,i} = \bigcup^{p}_{l=1} y^{k}_{n,l}\cdot I(h_j(l)=i)$
    \ENDFOR
    \newline
    \STATE \textbf{Training (on server):}
    \FOR {Synchronization round $t= 1,2,\ldots,T$}
    \STATE Randomly select a set $K_t$ that includes $S$ out of $K$ clients to collect
    \FOR {selected device $k \in K_t$ (in parallel)}
    
    \FOR {hash function $j=1,2,...,R$ (in parallel)}  
    \STATE $\vw^{j,k}_{(t+1)}$= DeviceTrain($j, k; \vw^{j}_{(t)}$)
    \ENDFOR
    \ENDFOR
    \FOR {hash function $j=1,2,...,R$ (in parallel)}  
    \STATE Aggregate parameter: $\vw_{(t+1)}^{j} = \sum_{k \in K_t}\frac{1}{S} \vw^{j,k}_{(t+1)}$
    \ENDFOR
    \STATE pass $\vw^{j}_{(t+1)}$, $j=1,2,\ldots,R$, to all the local clients
    \ENDFOR
    \newline
    \STATE \textbf{DeviceTrain($j, k; \vw$):}
    \STATE Receive $\vw^{j}_{(t)}$ from server 
    \FOR {each local epoch $i=1,2,, \ldots, E$}
        \STATE Update the parameters (in parallel): $\vw^{j,k}_{(t+1)} = \text{Train}(\vx^{k}, \vz^{j,k}; \vw^{j}_{(t)})$ for $j=1,2,\ldots,R$
    \ENDFOR
    \STATE Pass $\vw^{j,k}_{(t+1)}$ to the server
\end{algorithmic}
\end{algorithm}

During the $t$-th synchronization round, FedMLH randomly picks a set of $S$ clients (line 10, algorithm~\ref{alg:algo2}). On each selected client, all the sub-models are trained in parallel for $E$ epochs (line 11-15, algorithm~\ref{alg:algo2}). Then, send the updated local parameters to the central server. To update the global parameters, for each sub-model, the central server average the corresponding model parameters collect (line 16-18, algorithm~\ref{alg:algo2}). Finally, the global parameters are shared to all the local clients for the training of next synchronization round.

\paragraph{Parallelizable between sub-models}
Since different sub-models are fully independent, during the training, we do not need to communicate any parameters between different sub-models on the same clients. Actually, FedMLH learns a federated model for each sub-model in parallel.

\section{Analysis of FedMLH}

For the federated extreme classification tasks, there exists two significant problems: 1) The class distribution is highly imbalanced (due to the large number of classes); 2) The class distribution diverges between different clients (In FedAvg, class distribution means the number positive instances of different classes and in FedMLH, it refers to the number of positive instances in different buckets). We found FedMLH is helpful to relieve both problems.

\subsection{FedMLH adjusts the imbalanced class distribution}
\label{sec:adjust_imbalance}

\begin{figure*}[ht]
\centering
\subfloat[positive instance distribution \label{fig:label_freq}]{\includegraphics[width=0.31\linewidth]{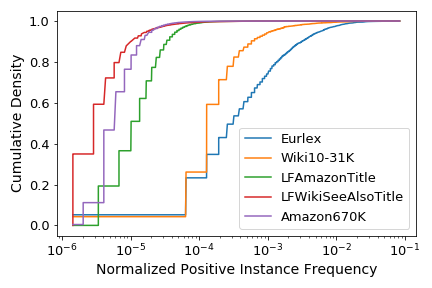}} \hfill
\subfloat[positive instance proportion distribution \label{fig:label_prop}]{\includegraphics[width=0.31\linewidth]{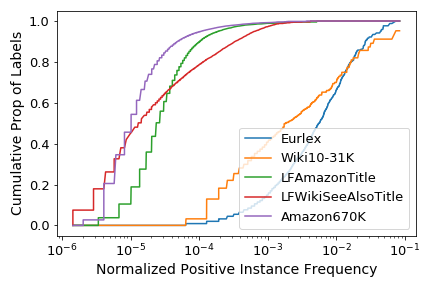}} \hfill
\subfloat[Non-iid data partition
\label{fig:non_iid}]{\includegraphics[width=0.38\linewidth]{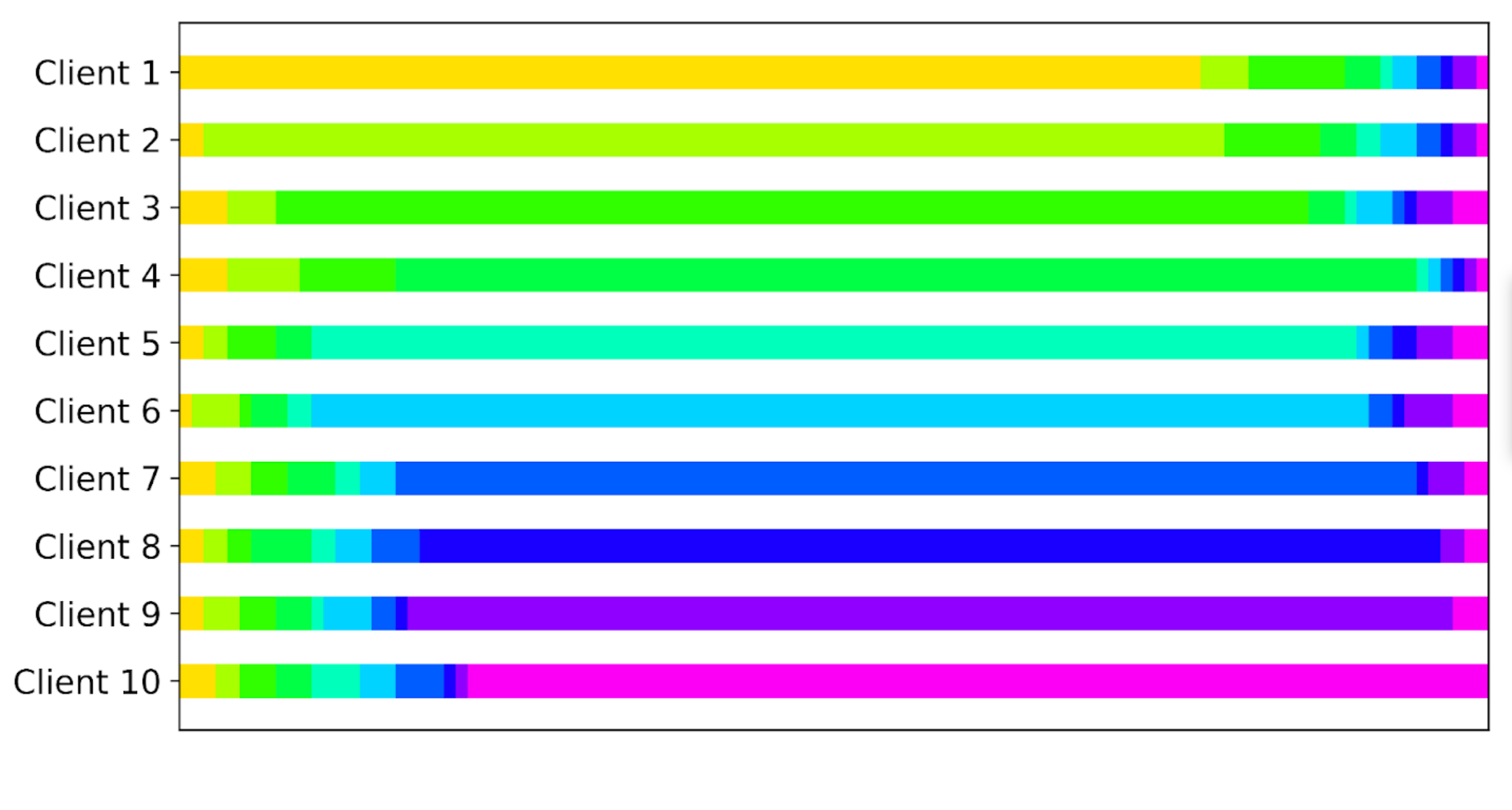}}\hfill
\caption{Distribution of (a) normalized positive instance frequency and (b) positive instance proportion. For each point $(x,y)$ on the line, (a) $y$ is the empirical proportion of $(\text{normalized positive instance frequency} \leq x)$; (b) proportion of positive instances contributed by the classes with normalized positive instance frequency less than $x$.
The distribution of positive instance frequency follows a power law in all the datasets. But infrequent classes also contribute a lot of positive instances. (c) non-iid data partition for extreme classification datasets in our setting. Each color represents the training samples associated with one frequent class. This bar plot shows the distribution of frequent class samples on local clients. $Y$ axis is the client id.}
\end{figure*}

When number of classes is huge, the number of positive instances are not evenly distributed among classes. In the real datasets, figure~\ref{fig:label_freq} suggests that only a proportion of classes have a lot of positive samples (called ``frequent classes''), while the majority of the classes just have a few samples (called ``infrequent classes''). On the other hand, these infrequent classes cannot be ignored. For example, figure~\ref{fig:label_prop} shows that for the ``LFAmazonTitle'' dataset, the classes with normalized label frequency (normalized positive instance frequency = \# of positive instances/sample size) less than $10^{-4}$ (less than $130$ positive instances) contributes about $70\%$ of positive instances. Therefore, if the classification model cannot predict the infrequent classes, it may miss more than $70\%$ of the positive instances, which is a huge loss. 

However, it is difficult to classify the infrequent classes in general due to the lack of positive instances. Theorem~\ref{theorem_1} suggests that if the class lacks enough positive instances, it is difficult to infer the distribution of positive samples on the embedding space, and both the centroid and radius of the positive sample cluster cannot well estimated.   

\begin{theorem}
\label{theorem_1}
Assume we observe $n$ i.i.d. samples of $(\vx, y)$, $(\vx_i, y_i)^n_{i=1}$, where $\vx_i \in \reals^d$ is the feature vector and $y_i \in \{0,1\}$ is the label of $\vx_i$. Let $n_1, n_0$ be the number of positive/negative samples of $y$ ($n_0 + n_1 = n$). Assume the model learns a embedding vector $m(\vx)$ which follows a mixed distribution of $m(\vx) \sim \pi f_1(m(\vx)|\vmu_1, \Sigma_1) + (1-\pi) f_0(m(\vx)|\mu_0, \Sigma_0)$, where $\vmu_0, \vmu_1$ are the mean of $f_0$ and $f_1$, and $\Sigma_0, \Sigma_1$ are the variance of $f_0$ and $f_1$. $\pi$ is the prior probability of $\PR{y=1}$. Thus, $(m(\vx)|y=0) \sim f_0(m(\vx)|\vmu_0, \Sigma_0)$ and $(m(\vx)|y=1) \sim f_1(m(\vx)|\vmu_1, \Sigma_1)$. Assume the Fisher information of $\vmu_0$, $\vmu_1$, $\Sigma_0$ and $\Sigma_1$ are bounded,
then for all of their unbiased estimators, we have
$\MSE{\hat{\vmu}_0(i)}, \MSE{\hat{\Sigma}_0(j,k)} \geq O(\frac{1}{n_0})$
and 
$\MSE{\hat{\vmu}_1(i)}, \MSE{\hat{\Sigma}_1(j,k)} \geq O(\frac{1}{n_1})$, where MSE is the mean square error. $\hat{\vmu}_0(i)$ is the $i$-th element of $\hat{\vmu}$, and $\hat{\Sigma}_0(j,k)$ is the $(j,k)$-th element of $\hat{\Sigma}_0$.
\end{theorem}

While for FedMLH, since the number of buckets $B$ is much smaller than the number of classes, multiple classes are merged into the same bucket. Thus, a bucket has much more positive instances on average compared to the positive instances in a class. Therefore, by theorem~\ref{theorem_1}, it is much easier for FedMLH to learn the bucket labels in each sub-model. 

In lemma~\ref{lemma:bucket_count}, we further quantify the change of positive instances. Since when the bucket size $B$ is not very small, $\frac{N_{lab}}{B^2}$ is almost negligible compared to the size of $\frac{1}{B}(N_{lab} -n_j)$. Hence, for any class $j$, the number of positive instances in its corresponding bucket increases by around $\frac{1}{B}(N_{lab}-n_j)$, which is a significant change especially for infrequent classes. For a example, if a class has $\frac{N_{lab}}{p}$ positive instances (equals to the average number of positive instances owned by a class), using the setup in our ``AMZtitle'' experiment, the corresponding bucket has \textbf{32} times more positive instances in expectation, which will significantly improves the estimation accuracy of positive sample cluster according to theorem~\ref{theorem_1}.

\begin{lemma}
\label{lemma:bucket_count}
Assume class $j$ is hashed into bucket $i$ in a hash table. Let $n_j$ be the number of positive instances in class $j$. Denote $N_{lab}$ as the total number of positive instances $N_{lab} = \sum_{j=1}^{p} n_j$. If the labels of different classes are independent to each other, then the expected number of positive instances in bucket $i$ is lower bounded by $\EX{B_i \mid h(j)=i} \geq n_j + \frac{1}{B}(N_{lab}-n_j) - \frac{N_{lab}}{B^2}$. 
\end{lemma}

However, the reduction of hash table size also faces some constrains. A typical constrain is the distinguishability of different classes. To ensure the classes are distinguishable, FedMLH uses multiple hash tables, and the size of each hash table cannot be too small. 
Lemma~\ref{lemma:bucket_size} gives requirement to ensure FedMLH is able to distinguish between different classes with high probability. If the size of the hash table is too small, there may exist some classes that are hashed into the same bucket in all the hash tables.
\begin{lemma}
\label{lemma:bucket_size}
Assume the $R$ hash functions used by FedMLH are independent to each. Given a $R \geq 1$, when $B \geq \left(\frac{p(p-1)}{2\delta}\right)^{1/R}$, then with probability $1-\delta$, there does not exist any two classes collide with each other in all the hash tables.
\end{lemma}

\subsection{FedMLH adjusts the non-iid class distribution}
Under the federated learning setup, the class distributions usually diverge a lot between different clients. In the federated extreme classification tasks, this problem becomes even more severe since divergence of class distributions increases with number of classes in general. However, FedMLH can relieve this problem in every sub-model. Theorem~\ref{theorem:divergence} suggests that the divergence of class distribution strictly decreases if we hash $p$ classes into less number of buckets. Moreover, as we hash into less buckets, the divergence monotone decreases in expectation. Therefore, FedMLH is helpful to adjust the non-iid class distributions and make the distributions more similar between different local clients. 

\begin{theorem}
\label{theorem:divergence}
Assume for each sample, only one class's label is positive. On client $k$, let $n^{(k)}_j = $ be the number of positive instances of class $j$. Then, $\vpi^{(k)} = (\pi^{(k)}_1, \pi^{(k)}_2, \ldots, \pi^{(k)}_p)$ is the proportion of positive instances of all the classes, where $\pi^{(k)}_j = \frac{n^{(k)}_j}{\sum_j n^{(k)}_j}$ is the proportion of positive instances of class $j$ $(\pi^{(k)}_j > 0)$. 
FedMLH hashes the $p$ classes into $B$ buckets, and on the $k$-th client, the proportions of positive instances of different buckets are $\vomega^{(k)} = (\omega^{(k)}_1, \omega^{(k)}_2, \ldots, \omega^{(k)}_B)$, where $\sum_{j=1}^{B} \omega^{(k)}_j =1$ and $\omega^{(k)}_j > 0$. Then, for any two clients $a$ and $b$, the Kullback–Leibler (KL) divergence between $\vomega^{(a)}$ and $\vomega^{(b)}$ is always smaller than that between $\vpi^{(a)}$ and $\vpi^{(b)}$.
\[
D_{KL}(\vomega^{(a)}, \vomega^{(b)}) < D_{KL}(\vpi^{(a)}, \vpi^{(b)})
\]
where $D_{KL}(\vpi^{(a)}, \vpi^{(b)}) = \sum_{i=1}^{p} \pi^{(a)}_i \log\frac{\pi^{(a)}_i}{\pi^{(b)}_i}$. 
\end{theorem}

\section{Experiments and Results}

We perform experiments to evaluate the performance of FedMLH on four different large scale extreme classification datasets, including EURLex-4K~\cite{mencia2008efficient} (Eurlex), Wiki10-31K~\cite{zubiaga2012enhancing} (Wiki31), LF-AmazonTitle-131K~\cite{mcauley2013hidden} (AMZtitle) and LF-WikiSeeAlsoTitles-320K~\cite{Bhatia16} (Wikititle). These datasets focus on potentially important application areas of federated learning with user generated data (NLP and recommendation system). The details of the four datasets are listed in Table~\ref{label_number}. Since the input features are sparse for most the extreme classification datasets, feature hashing is widely used to reduce the memory cost. Here, we also use feature hashing to reduce the feature dimension (Table~\ref{label_number} shows the hashed feature dimension). For training both baseline and FedMLH we use the same cluster of NVIDIA P100 gpus.

\paragraph{Baselines:} To evaluate our method, we compare FedMLH to the FedAvg algorithm. Both algorithm use the same MLP network (with two hidden layers) for each dataset, besides the last fully connect layer (FedMLH has less output). For different datasets, we vary the number of hash tables/sub-models ($R$) and number of buckets ($B$) used in each hash table (see Table~\ref{tab:bucket_size}). 

\paragraph{Non-iid data partition}

We manually partition the training samples to ensure the data on different local clients are non-iid distributed. Since the class distribution is highly imbalanced and most of the samples have at least one positive instances among the frequent classes, we try to partition the samples with frequent classes unevenly and make sure that the frequent classes on different local workers are distinct. In detail, for a frequent class $j$, we collect the training samples whose label of class $j$ is positive, denoted by $D^{(j)}$ ($D^{(j)} = \{(\vx_i, \vy_i): y_{ij} = 1\}$ where $y_{ij}$ is class $j$'s label of sample $i$). Then, we randomly pick a local client $k$, and assign $D^{(j)}$ to client $k$ (Figure \ref{fig:non_iid}). By this approach, different local clients have totally different frequent classes samples, thus have non-iid distributed data(\ref{summary_fig_2}). Since most of the samples have multiple labels, it is possible that $D^{(j)}$ and $D^{(l)}$ have non-empty intersections ($D^{(j)} \cap D^{(l)} = \{(\vx_i, \vy_i): y_{ij} = 1 \quad \text{and} \quad y_{il}=1 \}$). Therefore, samples with more than one positive instances among frequent class are assigned to multiple clients.

\begin{table}[ht]
\centering
\begin{tabular}{cccccc}
\toprule
    & Eurlex & Wiki31 & AMZtitle & Wikititle \\ \midrule
    $d$ & $5,000$ & $101,938$ & $40,000$ & $40,000$ \\
    $\tilde{d}$ & $300$ & $5,000$  & $5,000$ & $10,000$ \\ 
    $p$ & $3,993$ & $30,938$ & $131,073$ & $312,330$ \\
    $N$ & $15,539$ & $14,146$ & $294,805$ & $693,082$ \\
 \bottomrule
\end{tabular}
\caption{The statistics of feature dimension $d$, feature hashing dimension $\tilde{d}$, number of classes $p$ and number of training samples $N$.}
\label{label_number}
\end{table}

\begin{table}[ht]
\centering
\begin{tabular}{cccccc}
\toprule
    & Eurlex & Wiki31 & AMZtitle & Wikititle \\ \midrule
    $R$ & $4$ & $4$ & $4$ & $8$ \\
    $B$ & $250$ & $1,000$ & $4,000$ & $5,000$\\
 \bottomrule
\end{tabular}
\caption{Number of hash tables/sub-models ($R$) and buckets ($B$) used by FedMLH.}
\label{tab:bucket_size}
\end{table}

\paragraph{FL setups \& training details}

Our experiment includes 10 local clients, and during synchronization, we randomly pick 4 local clients to share the model parameters with central server. The models are trained for $70$ synchronization rounds, and each synchronization rounds contains $5$ epochs. 
We also apply early stopping on both baselines to achieve better accuracy and prevent overfitting. 
Due to the large input dimensions, we also perform feature hashing to all the datasets, and both baselines are run on the feature hashed data.

\begin{table*}[ht]
\centering
\begin{tabular}{ccccccc}
\hline
       &    & Eurlex & Wiki31 & AMZtitle & Wikititle  \\ \hline
       & @1 & $59.3\% (\textbf{+9.0\%})$      & $81.7\% (+1.1\%)$      & $18.3\% (+2.1\%)$        & $12.41\% (+3.0\%)$   \\
FedMLH & @3 & $45.6\% (\textbf{+7.3\%})$      & $63.4\% (+6.5\%)$      & $18.7\% (+1.3\%)$        & $11.89\% (+3.0\%)$  \\
       & @5 & $38.4\% (+5.1\%)$      & $52.1\% (\textbf{+6.3\%})$      & $20.4\% (+1.1\%)$        & $13.01\% (+3.4\%)$  \\ \hline
       & @1 & $50.3\%$      & $80.6\%$      & $16.2\%$        & $9.43\%$ \\
FedAvg & @3 & $38.3\%$      & $56.9\%$      & $17.4\%$        & $8.94\%$ \\
       & @5 & $33.3\%$      & $45.8\%$      & $19.3\%$        & $9.59\%$  \\ \hline
\end{tabular}
\caption{Top 1, 3 and 5 prediction accuracy of FedMLH and FedAvg.}
\label{accuracy}
\end{table*}

\begin{figure*}[ht]
\centering
\subfloat[Total Classes\label{fig:eurl_synchronization}]{\includegraphics[width=0.33\textwidth]{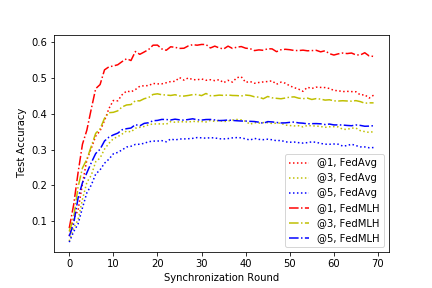}} \hfill
\subfloat[Frequent Classes\label{fig:eurl_freq}]{\includegraphics[width=0.33\textwidth]{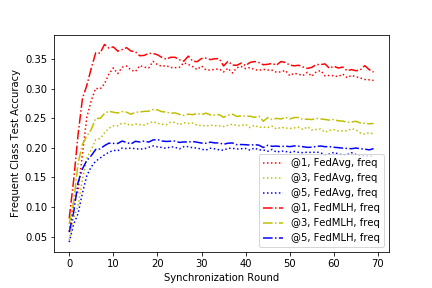}} \hfill
\subfloat[Infrequent Classes\label{fig:eurl_tail}]{\includegraphics[width=0.33\textwidth]{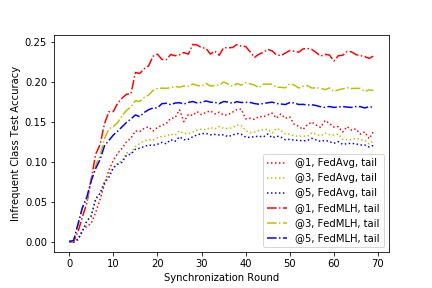}}
\caption{FedMLH vs FedAvg test accuracy of total classes, frequent classes and infrequent classes of Eurlex dataset in synchronization rounds. @1, @3, @5 means the precision at top 1, 3 and 5 selected classes}
\label{fig:eurl_acc}
\end{figure*}

\paragraph{Performance metrics}

Since both baselines multi-label classification models, traditional accuracy does not apply here. Instead, we evaluate the prediction accuracy using the top $1$, $3$ and $5$ accuracy. The top-$k$ accuracy is measured by the precision of the top $k$ classes with largest predicted log-probability, which defined as follows:
\[
\text{top $k$ accuracy} = \sum^N_{i=1} \frac{|P_k(\vx_i) \cap S_{\vy_i}(\vx_i)|}{Nk},
\]
where the $P_k(\vx_i)$ is the set of top $k$ classes with largest predicted probability of sample $i$, and $S_{\vy_i}(\vx_i)$ is the set of classes whose labels of sample $i$ are positive ($y_{ij}=1$).

\paragraph{Communication cost} Communication cost is another important concern of FL algorithms. We compare the communication volume of both baselines. The communication volume is defined as the size of the model parameters (in bytes) communicated between local clients and central server during the training. Here we measure the communication volume until the model achieves the best accuracy (the average of top $1$, $3$ and $5$ accuracy).

\subsection{Evaluation Results}

We evaluate the performance of FedMLH in terms of the predicton accuracy, communication cost, model size, convergence rate and training time.

\paragraph{Prediction Accuracy} Table~\ref{accuracy} suggests that compared to FedAvg, FedMLH significantly improves prediction accuracy in all the experiments. Especially for the EURLex-4K experiment, the top $1$, $3$ and $5$ accuracy are improved by $\textbf{9\%}$, $7.3\%$ and $5.1\%$ respectively, which is a significant boost. Moreover, in the LF-AmazonTitle-131K experiment, although the absolute accuracy improvement is not as high as that in the EURLex-4K experiment, considering the low baseline accuracy of FedAvg algorithm, the relatively accuracy improvements are even more remarkable (relative accuracy improvement is defined as: absolute accuracy improvement/baseline accuracy), which reach $\textbf{31.8\%}$, $\textbf{33.6\%}$ and $\textbf{35.5\%}$ for the top $1$, $3$ and $5$ accuracy respectively. 

We further evaluate the prediction accuracy of the frequent and infrequent classes. The top-$k$ frequent/infrequent class accuracy is defined as, \# of correctly predicted frequent/infrequent class labels/$k$ (sum of top-$k$ frequent class accuracy and infrequent class accuracy is the overall top-$k$ accuracy defined in ``Performance metrics''). We find most of the accuracy improvement comes from the infrequent class accuracy. In the EURLex-4K experiment, the top-$k$ frequent class accuracy of both baselines are almost the same (figure~\ref{fig:eurl_acc}). But FedMLH significantly outperforms FedAvg in terms of the infrequent class accuracy. This difference may be contributed to the adjustment of imbalanced class distribution accomplished by FedMLH (see section~\ref{sec:adjust_imbalance}).

\paragraph{Communication cost} During every synchronization round, FedMLH and FedAvg need to synchronize the model parameters between the local clients and central server. And the communication cost is a big bottleneck of FL algorithm.
We compute the communication volume of both baselines, and table~\ref{tab:comm} suggests that FedMLH significantly reduces the communication volume in all the experiments. Especially for the AMZtitle experiment, to reach the best accuracy, FedMLH achieves \textbf{18.75}$\times$ reduction of the communication volume, which will significantly reduce the communication time.

\begin{figure*}[ht]
\centering
\subfloat[Eurlex-4K\label{fig:eurl_communication}]{\includegraphics[width=0.45\textwidth]{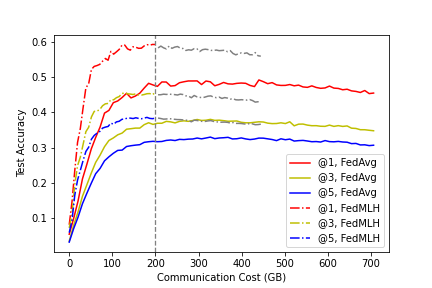}} \hfill
\subfloat[LF-AmazonTitle131K\label{fig:AMZtitle_communication}]{\includegraphics[width=0.45\textwidth]{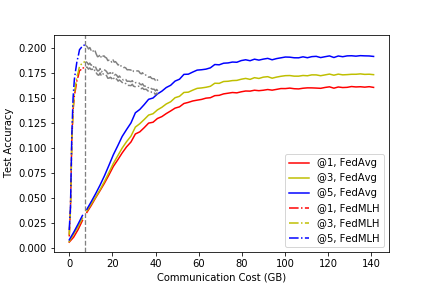}}\hfill
\subfloat[Wiki10-31K\label{fig:Wiki_communication}]{\includegraphics[width=0.45\textwidth]{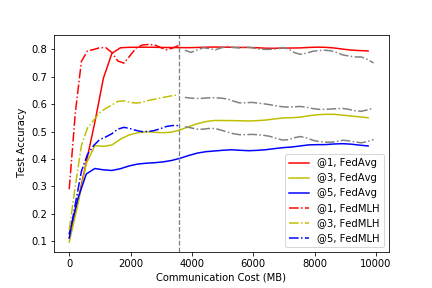}}\hfill
\subfloat[LF-WikiSeeAlsoTitles-320K\label{fig:Wikititle_communication}]{\includegraphics[width=0.45\textwidth]{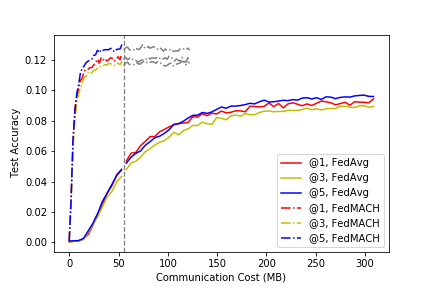}}
\caption{Test accuracy vs total communication volume transmitted by all workers. @1, @3, @5 means the precision at top 1, 3 and 5 selected classes. The vertical gray dash line indicates the place where we early stop the training of FedMLH.}
\label{fig:communication}
\end{figure*}

\begin{figure*}[ht]
\centering
\subfloat[Eurlex-4K: sensitivity to hash table size \label{fig:eurl_B}]{\includegraphics[width=0.45\textwidth]{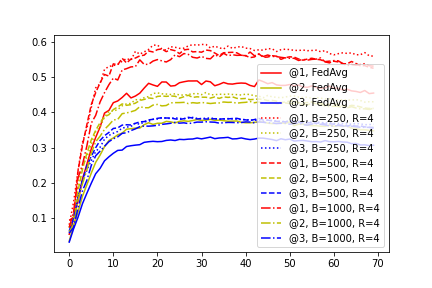}} \hfill
\subfloat[Eurlex-4K: sensitivity to number of hash tables \label{fig:eurl_R}]{\includegraphics[width=0.45\textwidth]{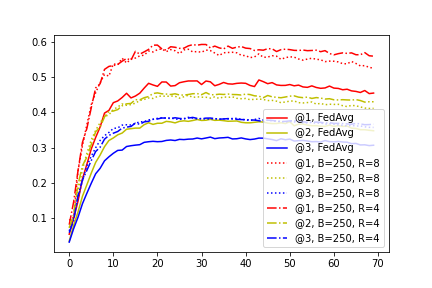}}\hfill
\subfloat[Wiki10-31K: sensitivity to hash table size \label{fig:Wiki_B}]{\includegraphics[width=0.45\textwidth]{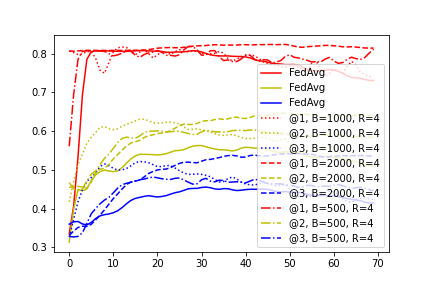}}\hfill
\subfloat[Wiki10-31K: sensitivity to number of hash tables \label{fig:Wiki_R}]{\includegraphics[width=0.45\textwidth]{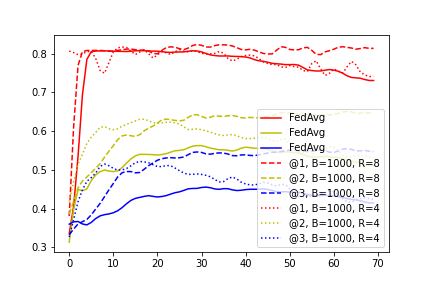}}
\caption{Test accuracy of FedMLH under different setups of hyper-parameters. @1, @3, @5 means the precision at top 1, 3 and 5 selected classes.}
\label{fig:hyperparameter}
\end{figure*}

\paragraph{Model size} FedMLH leverages label hashing to reduce the size of each sub-model by reducing the size of output layer. Although FedMLH requires multiple sub-models, the total model size is still reduced compared to that used by FedAvg. For example, in AMZtitle experiment, FedMLH reduces the model size from $0.51$GB to $0.15$GB, which also lowers the memory requirement of local computing devices.

\begin{table}[ht]
\centering
\resizebox{0.5\textwidth}{!}{%
\begin{tabular}{cccccccc}
\hline
        & Eurlex & Wiki31 & AMZtitle & Wikititle \\ \hline
FedMLH  & $199.7$Mb     & $3,572.8$Mb      & $7.2$Gb        & $53.4$Gb          \\
FedAvg  & $399.2$Mb     & $8,633.1$Mb      & $135.0$Gb        & $308.7$Gb        \\ 
CC Ratio       & $1.99\times$      & $2.41\times$      & $\textbf{18.75}\times$        & $5.78\times$    \\ \hline
\end{tabular}
}
\caption{Communication Volume of FedMLH and FedAvg to reach the best prediction accuracy. CC Ratio: Communication cost ratio of FedAvg over FedMLH.}
\label{tab:comm}
\end{table}

\begin{table}[ht]
\centering
\resizebox{0.5\textwidth}{!}{%
\begin{tabular}{ccccccc}
\hline
        & Eurlex & Wiki31 & AMZtitle & Wikititle \\ \hline
FedMLH  & $1.61$MB     & $49.62$MB      & $0.15$GB       & $0.48$GB    \\
FedAvg  & $2.56$MB    & $69.62$MB      & $0.51$GB       & $1.21$GB   \\ 
Memory Ratio       & $1.59\times$      & $1.40\times$      & $\textbf{3.40}\times$        & $2.52\times$     \\\hline
\end{tabular}
}
\caption{Model memory usage of FedMLH and FedAvg in each client. Memory Ratio is the memory cost ratio of FedAvg over FedMLH.}
\label{tab:memory}
\end{table}

\paragraph{Convergence rate} FedMLH not just decreases the model size, but also significantly speeds up the convergence rate in terms of the synchronization rounds (or training epochs). For example, in the AMZtitle experiment, FedMLH reduces the number of synchronization rounds from 66 rounds to only 12 rounds compared to FedAvg algorithm, which will significantly reduce the training time.

\paragraph{Local training time}                     

Since FedMLH reduces the model size, it is also beneficial to reduce the local training time. Table~\ref{local_time} measures the time to train a local synchronization round (5 epochs on a local client). Compared to FedAvg, FedMLH also has shorter local training time in all the experiments. 

\begin{table}[ht]
\centering
\begin{tabular}{ccccccc}
\hline
        & Eurlex & Wiki31 & AMZtitle & Wikititle \\ \hline
FedMLH  & $31$    & $18$      & $12$       & $28$   \\
FedAvg  & $39$    & $31$       & $66$       & $64$  \\ 
Rounds Ratio   & $1.25\times$    & $1.72\times$  & $\textbf{5.5}\times$    & $2.29\times$      \\ \hline
\end{tabular}
\caption{Number of synchronization rounds to reach optimal accuracy of FedMLH and FedAvg. Rounds ratio is the number of synchronization rounds of FedAvg over FedMLH.}
\label{table:conv}
\end{table}

\begin{table}[ht]
\centering
\begin{tabular}{ccccccc}
\hline
        & Eurlex & Wiki31 & AMZtitle & Wikititle  \\ \hline
FedMLH  & $4.67$s     & 5.44s      & $5.97$s       & $5.82$s    \\
FedAvg  & $5.38$s     & 5.73s      & $6.21$s       & $7.26$s    \\ 
Time Ratio       & $1.15\times$      & $1.05\times$      & $1.04\times$        & $1.24\times$         \\\hline
\end{tabular}
\caption{Wall clock time of FedML and FedAvg of each synchronization round. Time ratio: computation time of each synchronization round of FedAvg over FedMLH.}
\label{local_time}
\end{table}

\subsection{Hyper-parameter Tuning}

FedMLH includes two hyper-parameters, hash table size $B$ and  $R$ before running the experiments. A larger $B$ or $R$ leads to higher prediction accuracy. However, due to the memory constrain, we have to restrict the size of $B$ and $R$. In this section, we test the performance of FedMLH under different $B$ and $R$. First, we compare the sensitivity of FedMLH to the size of hash table size. Figure~\ref{fig:eurl_B} and \ref{fig:Wiki_B} suggest that the accuracy of FedMLH almost keeps the same when the number of hash tables doubles from $4$ to $8$. Hence, keep increasing the number of hash tables may not be helpful. 
From the memory perspective, a smaller $R$ is preferred. We also evaluated the effect of hash table size. Again, FedMLH is robust to the changes of hash table size. For the Wiki10-31K dataset, a larger hash table size could boost the top 3 and 5 accuracy by $5\%$. However, compared to FedAvg, FedMLH still significantly outperforms FedAvg even when we reduces the hash table size from $1,000$ (in the previous experiment) to $500$.

\section{Related Works}

\paragraph{Federated Learning in non-iid data} Federated learning has significant degraded performance in non iid datasets, first empirically observed by \cite{zhao2018federated}. In \cite{karimireddy2020scaffold}, FedAvg is shown to suffer from so called client-drift. Several analysis of FedAvg bound this drift by assuming bounded gradients \cite{wang2019adaptive, yu2019parallel} while some view it as additional noise \cite{khaled2020tighter}. Some work proposes solutions to such problem such as using variance reduction \cite{liang2019variance}, or adding regularization to the local worker training \cite{li2018federated}. But none of the above considered federated learning on extreme classification with imbalanced and they do not provide convergence speedup at the same time of reaching higher accuracy. 

\paragraph{Extreme classification} As many important real life tasks can be modeled as the extreme classification, it becomes one of the most important area of research. Several papers explore the extreme classification in centralized training scenario \cite{choromanska2014logarithmic,prabhu2018parabel,hsu2009multi,medini2019extreme}. But FedMLH is the first to explore and analyze extreme classification in highly imbalanced and non iid data distribution in federated learning.

\section{Conclusion}

In this project, we propose FedMLH for efficient federated extreme classification tasks. We demonstrate that our algorithm significantly reduces the communication cost, improves the classification accuracy and speeds up the training time, especially on the large scale extreme classification datasets. We envision that FedMLH will be widely implemented in different fields like the recommendation system.

\bibliography{reference}
\bibliographystyle{mlsys2022}




\end{document}